## Title:

*Investigation of intelligent barbell squat coaching system based on computer vision and machine learning.*

## Author Information

### Author names:

*Yinq-Rong Chern[1], Yuhao Lee[1], Hsiao-Ching Lin[1], Guan-Ting Chen[1], Ying-Hsien Chen[1], Fu-Sung Lin[1], Chih-Yao Chuang[2], Jenn-Jier James Lien[2] and Chih-Hsien Huang[*,1]*

### Affiliations:

*[1] Department of electrical engineering, National Cheng Kung University, Tainan 70101, Taiwan.*

*[2] Institute of physical education, health, and leisure, National Cheng Kung University, Tainan 70101, Taiwan.*

*[3] Department of computer science and information engineering, National Cheng Kung University, Tainan 70101, Taiwan.*

### Corresponding author:

*[*] Corresponding author: Chih-Hsien Huang (e-mail: chihhsien_h@mail.ncku.edu.tw), department of electrical engineering, National Cheng Kung University, Tainan 70101, Taiwan.*

# Abstract

**Purpose:** Research has revealed that strength training can reduce the incidence of chronic diseases and physical deterioration at any age. Therefore, having a movement diagnostic system is crucial for training alone. Hence, this study developed an artificial intelligence and computer vision-based barbell squat coaching system with a real-time mode that immediately diagnoses the issue and provides feedback after each squat. In addition, a replay mode allows users to examine their previous squats and check their comments. Initially, four primary characteristics of the barbell squat were identified: body joint angles, dorsiflexion, the ratio of knee-to-hip movement, and barbell stability. **Methods:** We collect 8,151 squats from 77 participants, categorizing them as good squats and six issues. Then, we trained the diagnosis models with three machine-learning architectures. Furthermore, this research applied the SHapley Additive exPlanations (SHAP) method to enhance the accuracy of issue prediction and reduce the computation time by feature selection. **Results:** The F1 score of the six issues reached 86.86%, 69.01%, 77.42%, 90.74%, 95.83%, and 100%. Each squat diagnosis took less than 0.5 seconds. Finally, this study examined the efficacy of the proposed system with two groups of participants trained with and without the system. Subsequently, participants trained with the system exhibited substantial improvements in their squat technique, as assessed both by the system itself and by a professional weightlifting coach. **Conclusion:** This is a comprehensive study that integrates artificial intelligence, computer vision and multivariable processing technologies, aimed at building a real-time, user-friendly barbell squat feedback and training system.

*Key words*—Computer vision, Intelligent sensors, Intelligent systems, Squat, Strength training

# 1 INTRODUCTION

The global population has recently been aging at a rate "without parallel in the history of humanity" [1] Based on the World Health Organization's definition, individuals aged 65 and older are typically categorized as elderly. The global elderly population was 6.87%, 7.55%, and 9.3% in 2000, 2010, and 2020, respectively [2]. As the human body ages, chronic diseases, physical decline, and cognitive impairment may develop [3, 4, 5, 6, 7]. These issues would impact society, such as the medical and care resources shortage.

Recent studies have consistently demonstrated that athletic ability significantly impacts the quality of life in older adults and is closely correlated with muscle mass [8, 9, 10, 11, 12]. Accordingly, engaging in strength training at any age is crucial for mitigating the effects of aging. In addition, encouraging strength training among individuals could potentially reduce the demand for healthcare resources and enhance societal productivity.

However, some issues will hinder the effectiveness of strength training for beginners because of immature techniques and physical capabilities. One limitation is the reduced ability to adjust posture effectively during exercise due to insufficient muscle mass and joint stability [13]. Another limitation is the increased risk of injury during training due to the gradual deterioration of ligaments and tendons [14, 15]. In addition, adhering to a long-term exercise regimen can be particularly difficult for beginners owing to a lack of supervision, motivation, and self-confidence [16, 17, 18].

Consequently, novice strength training participants often benefit from the guidance of a coach to address the aforementioned challenges. Under the supervision of a coach, beginners can receive personalized instruction, ensuring proper form and minimizing the risk of injury. In addition, the presence and support of a coach can facilitate the development of sustainable exercise habits. However, the coaches could not always be with the learner, and practicing alone was unavoidable. Therefore, an intelligent artificial coaching system capable of assessing exercise form and providing real-time feedback to novice strength training participants would be a valuable tool.

Recent advancements in machine learning have enabled significant progress in image recognition. Several researchers have published findings on utilizing deep neural networks to evaluate motion quality [19, 20]. For instance, L. Grewe et al. developed a squats classifier [20]. It extracted the features in each video frame utilizing EfficientNet and diagnosed the squats via the long short-term memory (LSTM) network. The proposed neural network achieved an accuracy rate of 80% on the test set. However, evaluating a single squat took approximately 2 min, which was not ideal for providing instance feedback.

In case of computer vision have enabled the development of systems capable of tracking human positions and identifying key joints within each frame. By representing human postures utilizing joint coordinates, data requirements can be significantly reduced. Consequently, machine learning models can more efficiently analyze motion data based on joint locations [21, 22, 23, 24]. For example, Gupta et al. employed OpenPose to extract 18 body key points, such as the hip and knee joints. These features were utilized to identify daily activities, including sitting, walking, running, and standing, using the K-nearest neighbors algorithm and the support vector machine model [25]. These previous studies demonstrated the potential to enhance the computational efficiency of motion recognition by utilizing reduced data inputs. However, most research only utilize pose estimation to identify the activity performed by the participant instead of grading its quality.

This study proposes an intelligent coaching system utilizing computer vision and machine learning to address the aforementioned challenges. The system would record and segment videos during each set based on the exercise sequence. Subsequently, it would provide a rating for each movement, along with specific advice and links to relevant video demonstrations.

For instance, when a user performs ten squats as a set, the proposed system would first recognize the start and finish of the set. Subsequently, the system would segment the entire set into ten squat clips and provide immediate feedback for each squat, including an evaluation, specific advice, and a link to a relevant video demonstration. In addition, the clipped videos and rating results would be automatically documented for users to look back on. Here, the barbell squat was selected as the target exercise. As a full-body compound exercise, it offers numerous benefits for older adults seeking to build muscle.

The study outlined its methodology in Section 2.1, detailing the experimental setup and data collection process. Section 2.2 described the data preprocessing, feature selection, and labeling techniques. The training and feature selection of machine learning models were discussed in Section 2.3. Section 2.4 introduced the development and validation of the proposed intelligent barbell coaching system. The results of data collection and preprocessing were presented in Section 3.1. Subsequently, Section 3.2 analyzed the performance of the machine learning models during training, and Section 3.3 evaluated the impact of feature selection. The efficacy of the proposed system was discussed in Section 3.4. Finally, Chapter 4 concludes the paper.

# 2  RESEARCH METHOD

## 2.1 Recording System and Data Collection

This study developed a squat recording system utilizing an Azure Kinect DK camera to collect data for training machine learning models. As depicted in Fig. 1, participants performed squats with their toes aligned with a reference line, facing the camera with their left side. The camera was positioned 135.5 cm away from the participant, at a height of 97 cm to capture the entire body. To facilitate path detection, the barbell was marked with red tape. This study utilized the pose estimation algorithm provided by the SDKs of the Microsoft camera to calculate the essential features for evaluating barbell squat movements, as illustrated in Fig. 2.

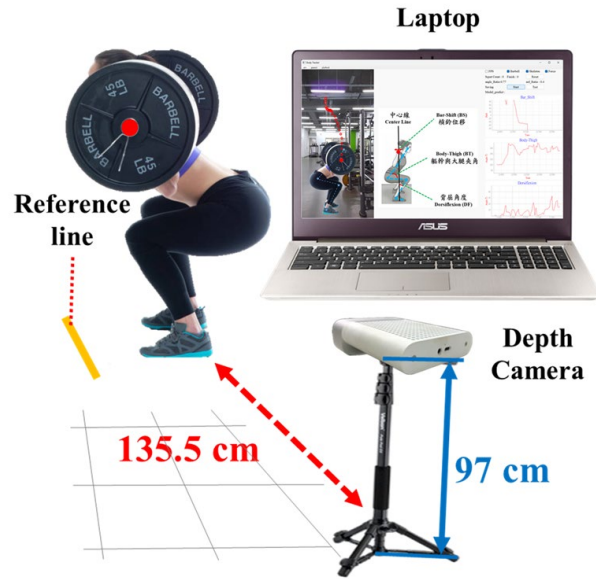

**Fig. 1 Schematic of squat recording system. Utilizing a depth camera to capture squat motion and collecting information with a user interface.**

The first feature was the angle of the body and thigh (body–thigh, BT, unit: degrees) calculated by the coordinates of the spine naval, pelvis, and knee. BT was an indicator of the mobility of the hip and bending of the upper body. In addition, this research utilized BT to facilitate the data recording and labeling. When the BT fell below 140°, the system would initiate video recording and feature extraction until BT returned to 140°. Subsequently, the system would automatically document each squat clip along with its corresponding features. The second feature, dorsiflexion (DF), measured the backward bending motion of the foot in degrees. This metric represents the ankle's mobility and the forward progression of the knee. However, a virtual vertical line was required to determine the dorsiflexion.

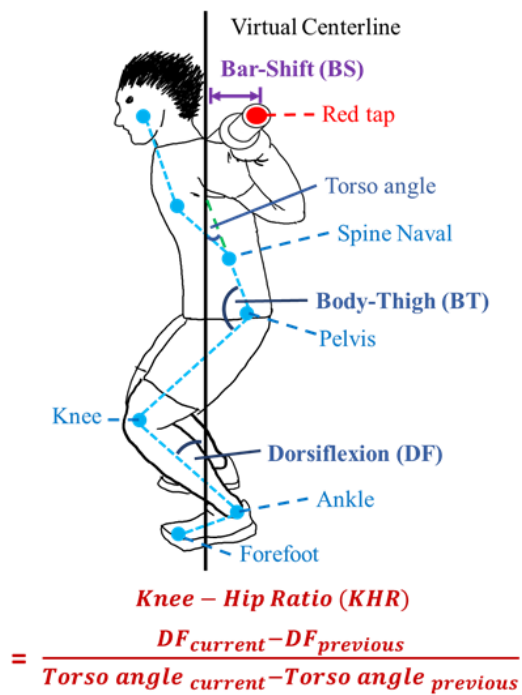

**Fig. 2 Schematic of four features collected here.**

Consequently, this study established a virtual centerline extending from the midpoint of the lateral coordinates of the ankle and forefoot. DF was then defined as the angle between this centerline and the tibia. The third feature considered was the ratio of forward and backward movements of the knee and hip within each frame (knee–hip ratio, KHR). It presented the proportion of power the quads and glutes generated during the squat. Fig. 2 illustrates that this study calculated KHR by dividing the current and previous frames' dorsiflexion and torso angle deviations. The torso angle was defined as the intersection angle between the extended line of the pelvis to the spine navel and the centerline. In addition, because the barbell's stability was a good indicator of body stability during squats, the barbell's shift (bar–shift, BS, unit: pixel) was collected as the fourth feature. The study estimated BS by utilizing OpenCV's color detection function to determine the coordinates of the red tape on the barbell and calculating its deviation from the virtual centerline. A total of 8502 squats were collected from 77 participants, including 42 males and 35 females with an average age of 50.

**2.2 Data Preprocessing and Labeling**

The pose estimation model might occasionally produce abnormal results during data collection because of lighting conditions, unexpected walk-in passersby, and participant's clothing. To ensure data quality, outliers in the collected dataset were identified and addressed. A reading was classified as an outlier if the BT exceeded 180°, the DF value exceeded 60°, the KHR surpassed 30, or the number of BS pixels exceeded 150. If a single squat contained multiple outliers, the entire squat was excluded from the analysis to prevent outliers from influencing the training results of machine learning models. Otherwise, outliers were replaced by interpolating the feature values of the preceding and subsequent frames. Here, 351 squat data were removed because they contained more than one outlier and 972 squat data were preserved after replacing the outliers with interpolation values. After that, the length of the features had to be unified because the time spent on each squat would be different, leading to variations in the number of frames. The average squat duration in this study was 45.3 frames with a standard deviation of 5.3 frames. After interpolation, all features were standardized to a size of 50. Finally, to facilitate the learning efficiency of machine learning models, this study applied three data processing methods, including variation (V), variation relative change (VRC), and Z score (Z). This work calculated V by subtracting the feature values of the current and the previous frame. In addition, VRC was the ratio of variation and the feature values of the last frame. The Z-score was calculated by subtracting the average feature value from each individual feature value and then dividing by the standard deviation. To ensure consistency in data length, the first value of each V and VRC vector was set to zero, as these features represent deviations. Consequently, this research created 12 datasets to train the machine learning models for motion quality diagnosis of barbell squats, as presented in Table 1.

**Table 1** The 12 datasets of this study were created via three data processing methods and four features

| Methods | Feature | Dataset |
|---|---|---|
| Variation (V) | Body-Thigh (BT) | BT-V |
|  |  | DF-V |
|  |  | KHR-V |
|  | Dorsiflexion (DF) | BS-V |
| Variation Relative Change (VRC) |  | BT-VRC |
|  |  | DF-VRC |
|  | Knee-Hip Ratio (KHR) | KHR-VRC |
|  |  | BS-VRC |
| Z score (Z) |  | BT-Z |
|  | Bar-Shift (BS) | DF-Z |
|  |  | KHR-Z |
|  |  | BS-Z |

According to Table 2, this study categorized the squats into seven labels and developed four models to recognize them. Label 1 represented a good squat, and Labels 2–7 were common issues during barbell squats. As illustrated in Fig. 3, Label 2 indicated an excessively shallow squat, where the femur was not parallel to the ground. Labels 3 and 4 represented posterior and anterior pelvic tilts, respectively, during the squat. Label 5 signified premature knee extension during the ascending phase of the squat relative to hip forward movement. Labels 6 and 7 denoted excessive hip or knee dominance during the descending phase of the squat.

**Table 2** The seven labels of this study

| Labels | Description | Number of data |
|---|---|---|
| 1 | Good squat | 2695 |
| 2 | Squat too shallow | 2280 |
| 3 | Posterior pelvic tilt | 839 |
| 4 | Anterior pelvic tilt | 491 |
| 5 | Hip rising too fast (ascending phase) | 1171 |
| 6 | Excessive hip dominant (descending phase) | 244 |
| 7 | Excessive knee dominant (descending phase) | 431 |
| | Total | 8151 |

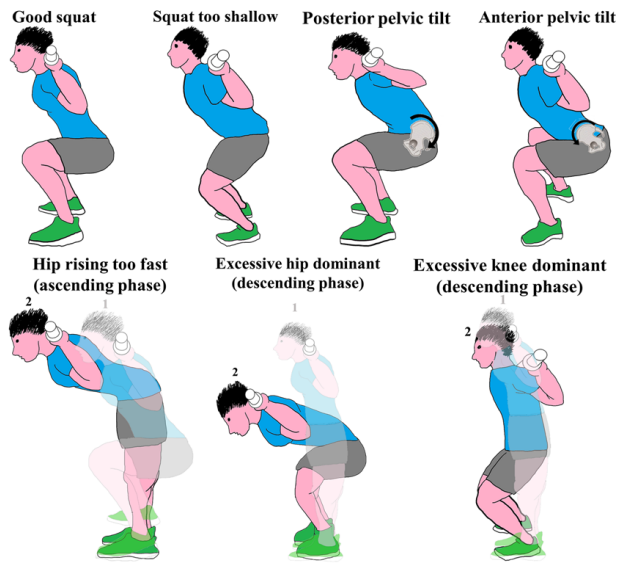

**Fig. 3.** Illustrations of the seven labels in this study. The numbers one and two in the graphs indicate the initial and following postures.

**2.3 Machine Learning Models for Squat Quality Diagnose**

As outlined in Table 3, this study developed four machine learning models to diagnose and grade barbell squats. The first model (Model A) was designed to assess squat depth sufficiency. Given that issues represented by Labels 3 to 7 inherently impact squat depth, this research independently collected data for Label 2 and trained the squat depth detection model solely on data from Labels 1 and 2. In contrast, the other models would utilize all the data for training. Given that posterior and anterior pelvic tilts do not typically occur simultaneously, Model B was designed to identify whether either of these issues, represented by Labels 3 or 4, was present. Furthermore, Model C was designed to identify excessive hip elevation during the ascending phase (Label 5). To address the issue of excessive hip or knee joint usage during the descending phase, as indicated by Labels 6 and 7, Model D was developed. Given that hip and knee dominance are mutually exclusive, these issues were treated separately. Table 3 provides a breakdown of the training classes and corresponding data quantities for each model. The dataset for each model was divided into training, validation, and testing sets in an 8:1:1 ratio.

This study tested three machine learning architectures commonly utilized to predict results from time series data. The first one was 1D convolutional neural networks (1D-CNN). This architecture mainly comprises convolution, pooling, and fully connected layers. The first layer was a 64×1 convolutional layer utilizing the ReLU as the activation function. The second layer was a max pooling layer, designed to preserve the most salient features following the convolution process. The third layer was a 128×1 convolutional layer. The fourth layer was a global average pooling layer, intended to emphasize the key characteristics of the data. The fifth layer was a fully connected layer with 512 neurons, employing the sigmoid activation function. The last layer was the output layer, which utilized Softmax activation for the multi-classification problem of Models B and D and the Sigmoid activation function for the binary classification problem of Models A and C. LSTM was the second architecture. The architecture comprised two LSTM layers, each with 100 LSTM units, followed by a dropout layer with a dropout rate of either 40% or 50%. The output layer of the LSTM model was identical to the 1D CNN model. The third architecture was a random forest with 100 estimators. The input number for each estimator was randomly determined but was constrained to be less than the square root of the feature number. For instance, if there were 12 features, the maximum feature number of an estimator would be three (the largest integer smaller than the square root of 12).

**Table 3** Four proposed squat diagnosis models with their data sizes and training conditions

| Model | Class 0 | Class 1 | Class 2 | Purpose of the Model |
|---|---|---|---|---|
| A | Label 1 (2695) | Label 2 (2280) |  | Squat too shallow |
| B | All ⊄ Label 3, 4 (6821) | Label 3 (839) | Label 4 (491) | Abnormal pelvic tilt |
| C | All ⊄ Label 5 (6980) | Label 5 (1171) |  | Hip rising too fast (ascending phase) |
| D | All ⊄ Label 6, 7 (7476) | Label 6 (244) | Label 7 (431) | Excessive hip/knee dominant squat (descending phase) |

This research initially trained the four models with the 12 datasets described in Section 2.2. However, it became apparent that each diagnostic model required distinct squat features for accurate predictions. Furthermore, model complexity was influenced by the number of inputs. Consequently, feature selection was deemed essential for enhancing both accuracy and computational efficiency. To streamline this process, the SHapley Additive exPlanations (SHAP) method was employed to refine dataset utilization for each model. SHAP values quantify the impact of each input on a model's output. This study calculated SHAP values for the 12 features corresponding to each model utilizing the SHAP kit provided by S. M. Lundberg and S.-I. Lee [26].

Because each input feature had a length of 50, 50 SHAP values were generated for each prediction result. However, it was not easy to analyze all their impact individually.

To facilitate comparison, the study averaged the SHAP values for each input feature. Theoretically, a higher SHAP value indicates a greater influence on the model's output. Moreover, an input feature is considered to contribute positively to the prediction if its SHAP value is positive, and vice versa. To identify models with both high accuracy and acceptable computational speed, this study retrained the four models utilizing two criteria. First, the input features were ranked by their SHAP values for each output, and the top 6 features were selected for each model. After that, based on the number of outputs, this study would pick intersected features from two or three outputs before training a novel model (intersected features). Second, this research would retrain the models after removing all the features with negative SHAP values corresponding to any outputs for each model (positive features).

This study evaluated the performance of squat diagnosis models utilizing the F1 score of each class, calculated as Equation (1).

$$F1 - Score = \frac{2TP}{2TP+FP+FN} \quad (1)$$

True positives (TP) refer to the number of times the model correctly identified a class that did happen. False positives (FP) and false negatives (FN) refer to the number of times the models identified a class that did not happen and failed to recognize a class that occurred, respectively.

**2.4 Development and Verification of Intelligent Coaching System**

Following model optimization, a grading function was developed to evaluate squat performance. As outlined in Table 4, the scoring system was based on point deduction, with a maximum score of 100 points. According to the opinions of powerlifting professionals, the deductions of issues corresponding to Labels 2–7 were 15, 20, 10, 10, 27.5, and 27.5 points.

The proposed intelligent squat coaching system, as depicted in Fig. 1, incorporated feature detection, grading, and replay functions. The data recording system described in Section 2.1 was modified to extract relevant features and assign a squat grade. As illustrated in Fig. 4, the system would record the barbell's initial position (typically on the squat rack) after the user initiated the session. Once the barbell moved horizontally more than ten pixels from its starting position, the system would begin tracking the BT angle. Subsequently, the system would automatically record each squat, providing immediate scoring results and issue identification. The system would automatically stop recording when the barbell returned to within ten pixels of its initial position. The system also included a replay function that allowed users to review their performance after each training set. The system would display scores, identified issues, and relevant suggestions, enabling users to select specific squats for review and adjust playback speed as needed.

Finally, this study examined the developed intelligent barbell squat coaching system with two groups of participants (five people in each group). The first group practiced barbell squats independently, while the second group trained with the proposed system. Both groups performed three sets of ten squats per week. Following training sessions, each participant completed an additional set of ten squats, which were recorded and graded by the proposed system to evaluate training progress. After four weeks, all 40 recordings from each participant in both groups were assessed by a powerlifting coach utilizing the same evaluation criteria outlined in Table 3 to validate the efficacy and accuracy of the proposed intelligent barbell squat coaching system.

**Table 4** Deduction points corresponding to each label

| Label | Description | Deduction point (Full points: 100) |
|---|---|---|
| 2 | Squat too shallow | -15 |
| 3 | Posterior pelvic tilt | -20 |
| 4 | Anterior pelvic tilt | -10 |
| 5 | Hip rising too fast (ascending phase) | -10 |
| 6 | Excessive hip dominant (descending phase) | -27.5 |
| 7 | Excessive knee dominant (descending phase) | -27.5 |

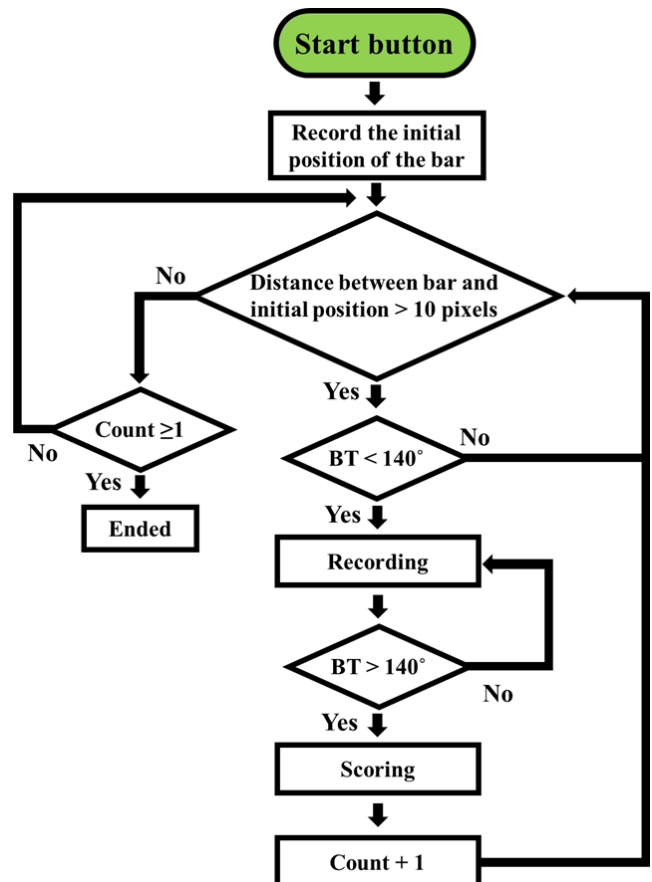

**Fig. 4.** Working flow of proposed intelligent barbell squat coaching system.

## 3 RESULT

### 3.1 Data Recording and Preprocessing

As illustrated in Fig. 5, the mean values of each feature at each data point were calculated after outlier removal and interpolation. When a participant performed a good squat (Label 1), BT and DF reached their minimum or maximum values simultaneously, while BS remained relatively unchanged. This suggests exceptional coordination between the torso and legs. In addition, the KHR behaved in an "S" shape because of the gradient of DF being larger than BT before closing to the squat's lowest position and becoming smoother than BT until the participant started to stand up. This was due to the knees bending rapidly at the onset of the squat and maintaining a relatively stable position at the bottom. In contrast, the torso ascended more rapidly at the beginning of the ascending phase to generate upward force. In contrast to the good squat (Label 1), the shallow squat (Label 2) exhibited lower BT and DF values at the lowest data point, indicating insufficient body and knee flexion. Accurate pelvic position was difficult to capture here because the participants wore sportswear with longer hems that covered their hips. Consequently, there was no significant difference in BT between the good squat (Label 1) and squats with posterior or anterior pelvic tilts (Labels 3 and 4). However, the DF of Label 3 was lower than Label 4 at the lowest position because the knees were more challenging to bend when the participant's pelvis tilted posterior compared to the anterior. As Label 5 indicates premature hip elevation during the ascending phase of the squat, it is logical to observe a delayed increase in BT from the lowest point in Label 5 compared to Label 1. Furthermore, Label 5's DF decreased earlier than Label 1's from the maximum point, suggesting that participants extended their knees to raise their hips at the beginning of the ascending phase. In addition, the BT would still increase, even if the participants focused on moving their hips first in the ascending phase. Therefore, BT may have exceeded 140°, prematurely terminating the recording and preventing the knees from fully extending. This could explain why the knee-to-hip ratio (KHR) value for Label 5 was more pronounced in the final period compared to the other labels. When the upper body bent forward faster than the knees during descending (Label 6), the BT reached its lowest value earlier than the others and increased because the knees kept bending forward, as illustrated in Fig. 5(a). As Fig. 5(b) indicates, the DF of Label 6 was small in the beginning and reached its highest value later than the others because the participants had to keep bending their knees to achieve the required depth. In addition, the BS was initially bigger than the others because the participants moved their hips backward more than shifting their knees forward in the descending phase, as Fig. 5(c) indicates. Owing to moving the knees more than the hips during the descending phase of the

10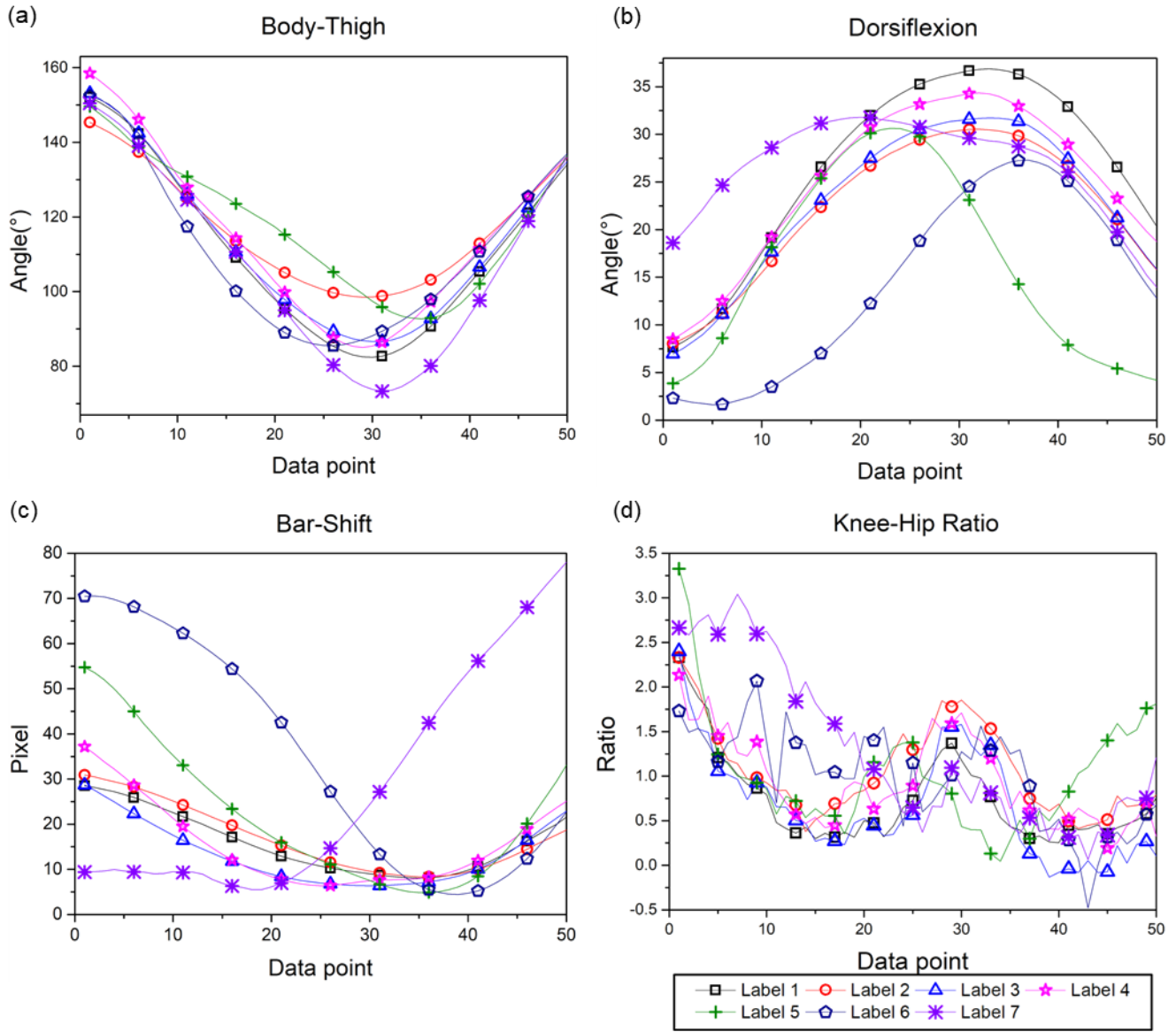

**Fig. 5.** Data preprocessed results of BT (a), dorsiflexion (b), BS (c), and KHR (d). The lines with hollow square, circle, triangle, star, cross, pentagon, and solid asterisk markers represent the mean values of Labels 1–7 at each data point, respectively.

squat (Label 7), Label 7 had greater DF and KHR values than the others before the participants paused at the bottom, as illustrated in Fig. 5(b) and (d). Furthermore, the barbell's trajectory was relatively straight owing to reduced upper body movement at the beginning of the descent, resulting in a flatter BS value. In the later stages of the descending phase, as the body's center of gravity shifted backward, many participants overcompensated by leaning forward excessively to avoid falling during the ascent. This resulted in a significantly larger BS value and a lower minimum BT value compared to other labels.

### 3.2 Machine Learning Models

As outlined in Section 2.3, this study evaluated three machine learning architectures by training each with three feature sets and subsequently calculating the F1 score for each class and the corresponding computation times. As depicted in Fig. 6, the F1 scores for 1D-CNN and LSTM models exceeded 80% across all classes in Models A, C, and D, while the random forest model consistently achieved F1 scores below 80%. Random forest was faster than the other two for diagnosing the same issues, and LSTM was more rapid than 1D-CNN. However, all the architectures could not accurately identify the posterior and anterior pelvic tilts. The accuracy of pelvic position detection may have been compromised by the sportswear worn by participants, which often has longer hems that obscure the hip region. In addition, features that introduce noise into the models' predictions can negatively impact the F1 score and should be excluded during model training. When trained with the reduced feature set, the models demonstrated significantly improved computational efficiency, with computation time inversely proportional to the number of features. Furthermore, the F1 score for most classes increased following feature selection with the SHAP method. As an example, the F1 score for Class 1 in Model A, trained with the 1D-CNN and LSTM architectures, increased from 78.6% and 83.81% to 83.93% and 85.05%, respectively. Although the F1 score of Classes 1 and 2 in Model B were lower than 80%, the F1 score of Class 2 after utilizing the intersected features rose from 74.7% to 77.42% for 1D-CNN and 45.71% to 70% for LSTM. Apart from this, the F1 score of Model B trained with 1D-CNN was significantly higher than others after utilizing intersected features, which reached 93.54%, 69.01%, and 77.42% for Classes 0, 1, and 2, respectively. In addition, the F1 score of the classes utilizing LSTM architecture in Model C improved after training with the intersected features, which could reach more than 90% (98.03% in Class 0 and 90.74% in Class 1). For Model D, the F1 score of Classes 1 and 2 utilizing LSTM with the intersected features increased from 93.62% to 95.83% and 98.95% to 100%, respectively. Because the SHAP values for all random forest models were negative, this study did not train this architecture with positive features. The training results of all features were adopted as the standard. The SHAP method could only slightly improve the score when the F1 score was relatively high (greater than 80%). However, when the F1 score was relatively low, the SHAP method could significantly improve the score while reducing the calculation time. When developing the intelligent squat coaching system, the primary criterion for selecting diagnostic models was the F1 score for Classes 1 and 2 (if available). However, computation time was also considered if multiple models exhibited comparable F1 scores. To provide immediate feedback, the proposed system attempted to display the grading result of each squat within 0.5 s.

Furthermore, the LSTM model trained with positive features and the 1D-CNN model trained with intersected features were selected as squat diagnosis Models A and B, respectively. Model C was similarly assigned to the LSTM model trained with intersected features, as it achieved the highest F1 score for Class 1 while maintaining acceptable computational efficiency. In Model D, the 1D-CNN model trained with positive features and the LSTM model trained with intersected features exhibited comparable F1 scores across all classes. Ultimately, the LSTM model was chosen for Model D because of its lower computational cost.

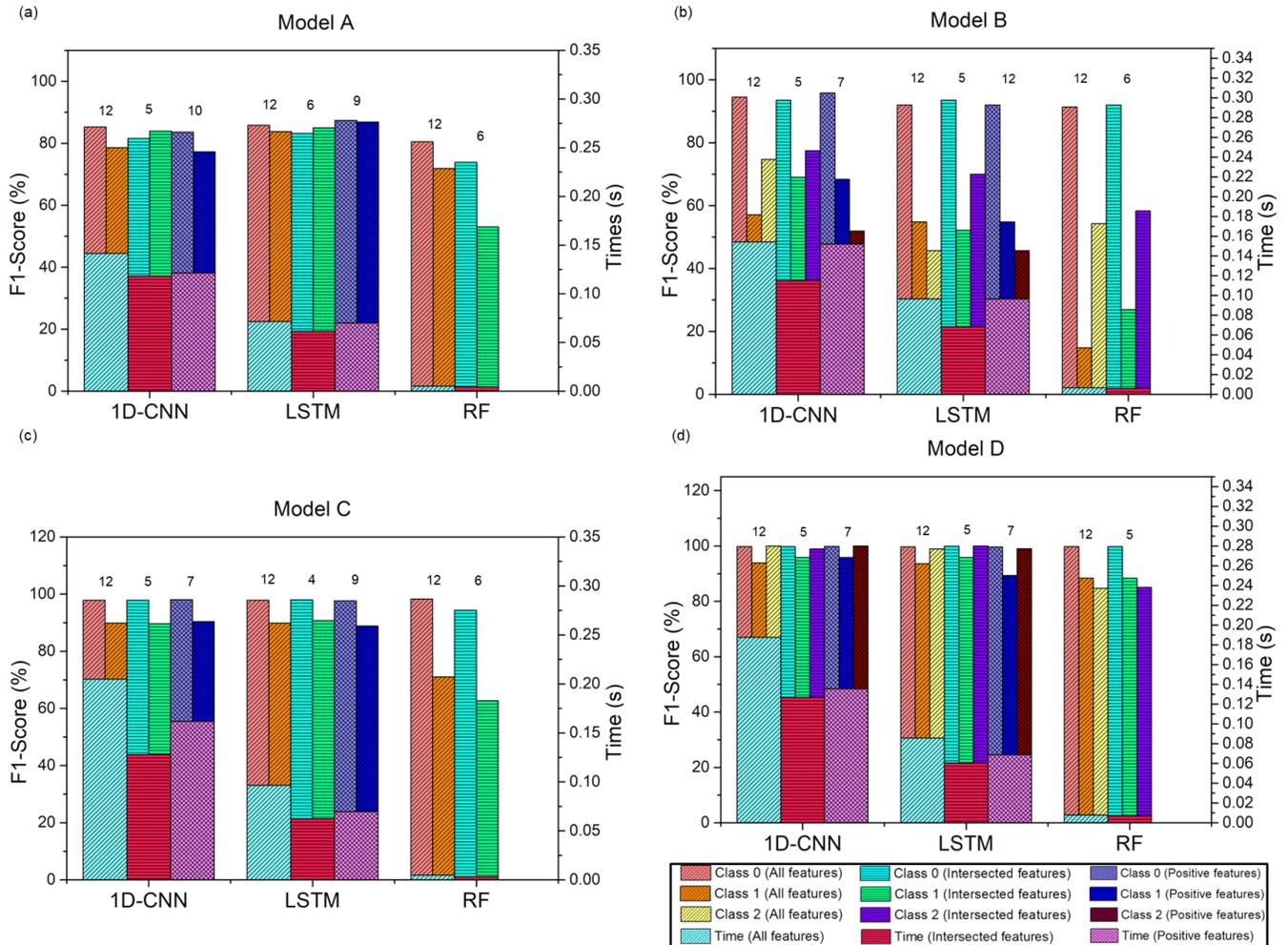

**Fig. 6.** F1 score and operation times of squat issue diagnosis models trained with each machine learning architecture are presented: (a) Model A. (b) Model B. (c) Model C. (d) Model D. The three feature combinations (all features, intersected features, and positive features) are represented by bars with diagonal lines, horizontal lines, and the grid, respectively. The feature number for training each model was on top of the bars.

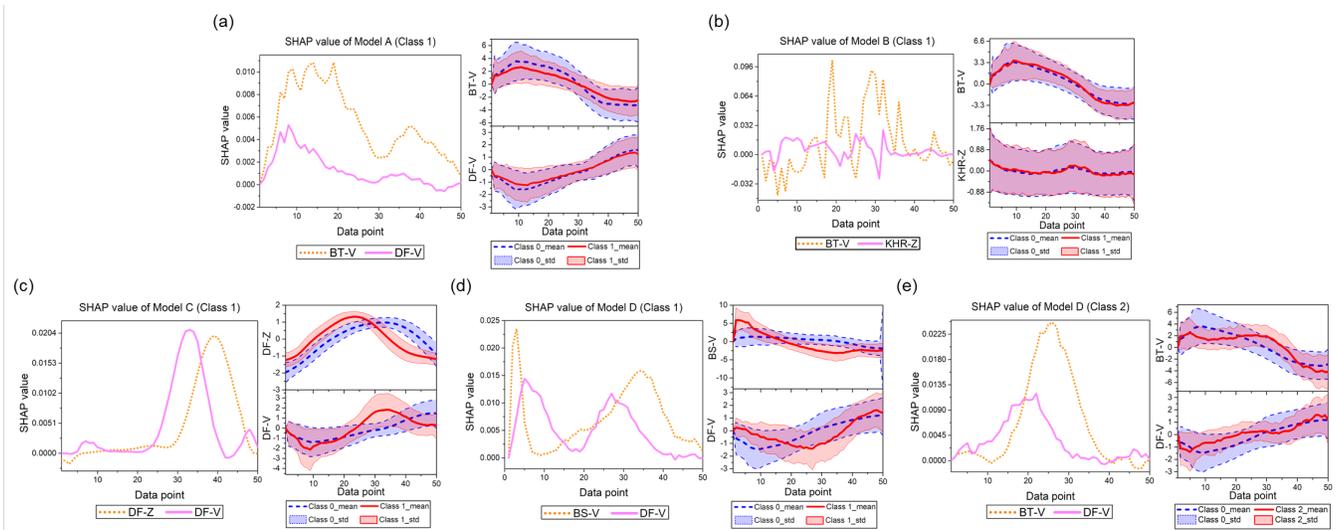

**Fig. 7.** Top two SHAP values from Class 1 of Model A (a), Model B (b), Model C (c), Model D (d), and Class 2 of Model D (e), along with their corresponding datasets. The dotted and solid lines represent the first and second highest SHAP values, respectively. In the right-hand graphs, the dashed and solid lines depict the mean values for Classes 0 and 1/2 of each dataset. The shaded region indicates the area of their mean ± standard deviations.

**3.3 SHAP Method**

As demonstrated in the previous section, the SHAP method effectively improved model F1 scores. To further validate its efficacy, this section explores the relationship between SHAP values of specific features and the identified issues in barbell squats. As illustrated in Fig. 7, this study plotted the SHAP values and the datasets of the two most significant features out of 12. First, Fig. 7(a) demonstrates that the SHAP values of BT-V and DF-V for Class 1 of Model A with LSTM architecture were higher in the descending phase. This suggests that the speed of body bending, particularly in the knees, is a critical factor in achieving sufficient depth at the lowest point. Furthermore, the SHAP values of BT-V were also higher in the ascending phase because the mobility of hip joints usually dominates the squat depth. In comparison to squats without sufficient depth (Class 1), good squats (Class 0) exhibited greater body flexion and ascended more rapidly, aligning with the periods of higher SHAP values.

Fig. 7(b) presents the datasets with the top two SHAP values (BT-V and KHR-Z) corresponding to Class 1 of Model B trained with 1D-CNN architecture for diagnosing posterior pelvic tilt (Label 3). Because the posterior pelvic tilt usually happened when the participants tried to squat down deeper after their hip joints were locked, the BT-V indicated a more apparent difference between Classes 1 and 0 (squats without posterior or anterior pelvic tilt) in this period. Furthermore, the SHAP value of BT-V exhibited a notable increase during this phase. Owing to the challenges associated with accurately detecting the spine navel through clothing, the F1 score for identifying posterior pelvic tilt was relatively low, as depicted in Fig. 6(b). Consequently, the SHAP method encountered limitations in identifying critical intervals within the datasets. For instance, the SHAP value of KHR-Z lacked a distinct prominent period, despite being the second most significant feature. Moreover, Fig. 7(b) illustrates the overlap between the KHR-Z values of Classes 1 and 0. Fig. 7(c) illustrates the datasets with the top two SHAP values (DF-Z and DF-V) for Model C, a 1D-CNN architecture trained to diagnose excessive hip elevation (Label 5). Since this issue typically occurs during the ascending phase, the SHAP values for DF-Z and DF-V are more prominent during this period. As previously mentioned, the recording process may terminate prematurely if the participant's hips ascend too quickly relative to their knees, leading to a higher peak in DF-Z for Class 1 compared to Class 0. This demonstrates the SHAP method's ability to effectively identify this crucial difference. Furthermore, because the participants relied on straightening their knees to stand up, the DF-V of Class 1 was also larger than Class 0 at the beginning of ascending, and this period's SHAP values of Class 1 also increased.

The SHAP values for BS-V and DF-V were the top two features for Model D, an LSTM architecture trained to identify excessive hip dominance during the descending phase (Class 1). As seen in Fig. 7(d), Class 1's BS-V and DF-V significantly differed from Class 0 in two stages.

The first stage of the squat involved minimal knee movement, requiring participants to descend primarily by leaning forward.

This resulted in a pronounced BS-V value and a relatively smaller DF-V value compared to squats without excessive hip dominance (Class 0). During the second stage, the knees began to bend, and the barbell quickly returned to the body's center, causing a reversal in the SHAP values of BS-V and DF-V. Although the BS-V and DF-V of Classes 0 and 1 did not change simultaneously, the SHAP values did respond to their differences in the corresponding periods. Fig. 7(e) illustrates the SHAP values of BT-V and DF-V for Class 2, representing excessive knee dominance during the descending phase. While the BT-V values for Classes 2 and 0 differed significantly in the early stages of the squat, the SHAP value for BT-V did not increase until participants began to ascend. This may be attributed to the sudden hip elevation performed by participants to relieve knee stress upon reaching the lowest position. The SHAP method may have interpreted this as a distinct behavior compared to squats where the knees are engaged without excessive hip involvement. Furthermore, the SHAP value for DF-V in Class 2 exhibited a significant increase after the participant's knees locked in the early stages of the squat. Fig. 7 demonstrates that the SHAP method could successfully identify the most significant difference between squats with and without the target issues in most cases. Moreover, the accuracy of the trained models was directly correlated with the magnitude of the SHAP values associated with the corresponding issues.

**3.4 Verification of Real-Time Coaching System**

Fig. 8 represents the user interface of the proposed intelligent barbell squat coaching system.

The user interface allowed for seamless switching between real-time feedback and playback modes utilizing the top-left corner tabs. In real-time feedback mode, the system displayed the current frame on the left and the live raw data of the four features in the middle. A graph in the lower-right corner provided visual representations of the four features. The control panel in the top-right corner displayed the squat count, current status, and prediction results for the most recent squat. As illustrated in Fig. 8(a), pressing the start button changes the status from "Start" to "Stop." When the participant lifts the barbell from the rack to the reference line, the status transitions from "Stop" to "Running." The system initiates frame recording when the BT falls below 140°, and the model prediction status changes to "Predicting." As demonstrated in Fig. 8(b), the system determines that a squat is complete (squat count increments by 1) when BT exceeds 140°. Furthermore, the model prediction status would display the diagnostic results, and the score would appear below the start/stop button. If the system detected multiple outliers within a single feature, the current squat would not be scored, and the model prediction status would change to "Data Error." Upon completing a set and returning the barbell to the rack, the button's status would revert from "Stop" to "Start."

As illustrated in Fig. 8(c), switching to playback mode in the top-left corner displayed the selected squat video in the left panel. The video control panel below allowed users to start, pause, and adjust playback speed. In the center of the playback mode screen, explanations and recommendations for the six squat issues were provided. On the right side of the screen, the recordings of the four features were displayed, with a slider on each graph that moved in synchronization with the current frame.

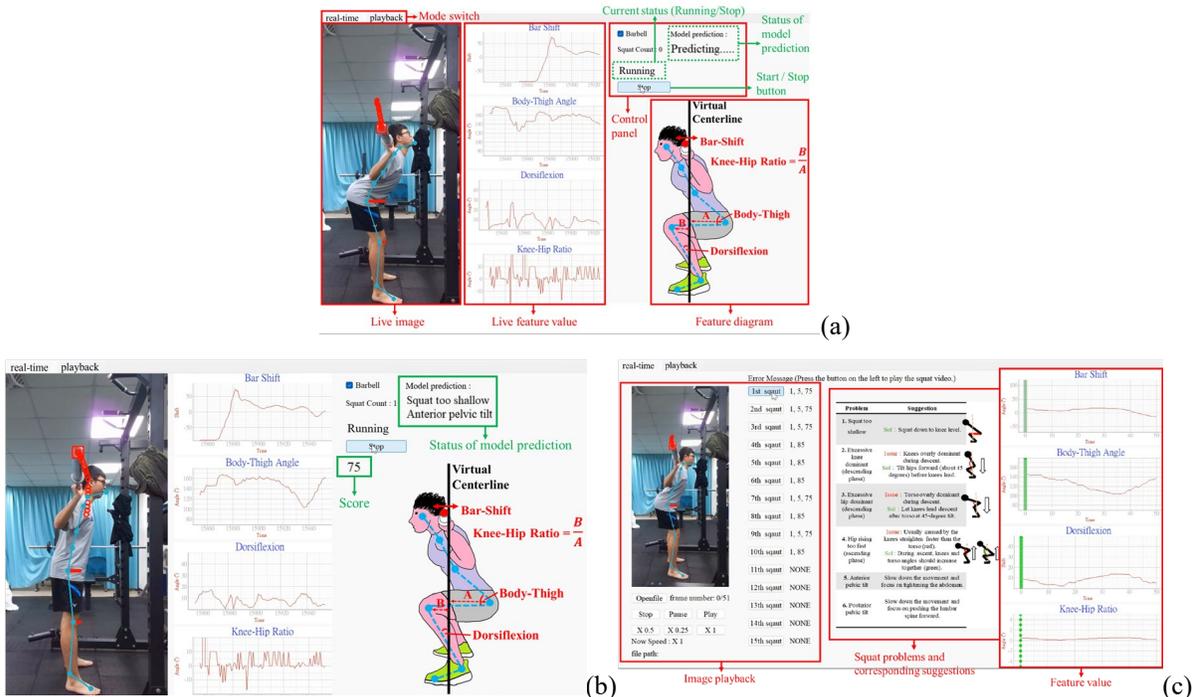

**Fig. 8.** User interface and functions of proposed intelligent barbell squat coaching system: (a) real-time mode while initiating squat, (b) real-time mode when finishing squat, and (c) playback mode for displaying recorded features and video of each squat.

Fig. 9 presents the efficacy of the proposed system. Fig. 9(a) and (b) present the proposed system and the weightlifting coach scoring the two groups of participants' squats before (Week 0) and after (Weeks 1–4) training, respectively. Participants who trained with the proposed system demonstrated significantly improved performance compared to those who trained independently during the final week. In addition, the participants who received immediate feedback from our system exhibited continued improvement week after week in both grading results most of the time. Participant 3 exhibited notable inconsistencies due to clothing choices. During the first three weeks, loose-fitting attire hindered the system's and coach's ability to accurately identify pelvic tilt issues, as the hemline often obscured the hip region. Hence, Participant 3 did not notice this issue, and the coach finally caught it in the Week 4 videos.

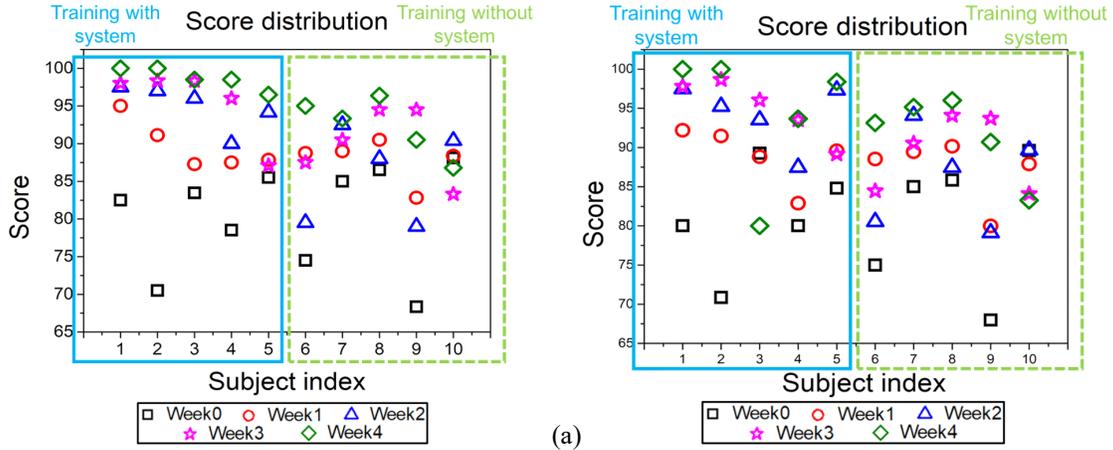

**Fig. 9.** (a) Recorded scores for participants training with (Participants 1–5) and without (Participants 6–10) proposed system. (b) Scores graded by powerlifting coach. Scores from Week 0 (before training) to Week 4 are represented by boxes, circles, triangles, stars, and diamonds, respectively.

As depicted in Fig. 10(a), the mean scores of participants training with the proposed system surpassed those of the control group after Week 1, with P-values falling below 0.05 after Week 3. This suggests that the intelligent coaching system effectively guided participants in mastering proper squat technique and achieving higher scores according to the established scoring algorithm. Fig. 10(b) presents the scores of the training records evaluated by a powerlifting coach. The mean scores of participants trained with the proposed system surpassed the others after Week 1, and the P-values also became significant after Week 2. However, unlike the results graded by the system, the P-values did not keep improving. Hence, the authors removed Participant 3, who had issues with clothing, and the P-values continued to improve in this situation. Therefore, the efficacy of the proposed coaching system should be considered significant, particularly when user attire does not hinder the accuracy of the pose estimation algorithm.

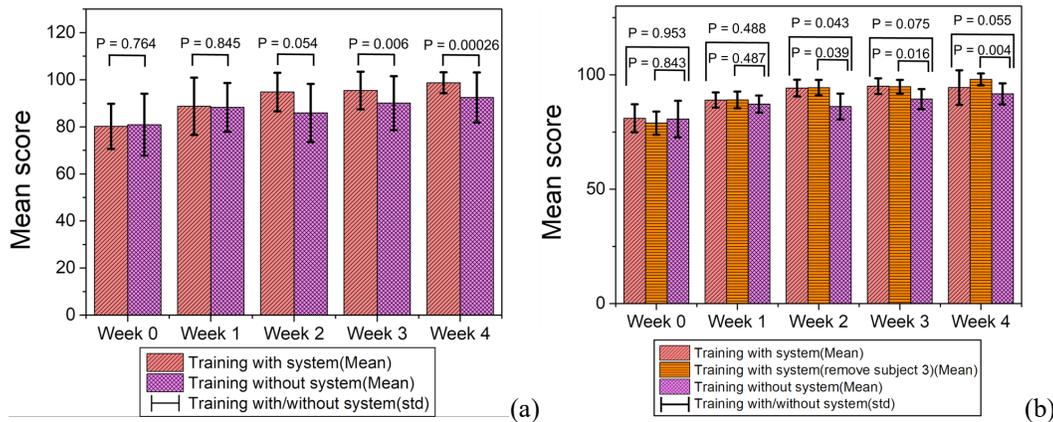

**Fig. 10.** Mean scores of groups trained with (diagonal line) and without (grid) proposed system for each week graded by proposed system (a) and powerlifting coach (b). P-values for weekly scores are on top of the bars.

Fig. 11 represents the confusion matrix of the squats diagnosed by the proposed system (predicted labels) and a powerlifting coach (actual labels). In addition, the bottom of Fig. 11 lists the F1 scores of the verification tests (Experiment) and the selected squat diagnosis models (ML models). The F1 scores of the experimental results closely aligned with the testing sets, except for the pelvic tilting issues (Labels 3 and 4). As mentioned previously, participants' outfits affected the detection of hip joints, making it difficult for the system to accurately identify the posterior pelvic tilt issue (Label 3) and misjudged many of them as good squats (Label 1). In addition, the F1 score for Label 4 was poor because of anterior pelvic tilt only occurring five times in this verification experiment.

For issues not directly related to hip joints, the F1 scores of this experiment closely aligned with the testing sets, even if such instances were infrequent, such as excessive hip elevation during the ascending phase (Label 5). Based on these findings, the proposed system can be considered reliable and valid for diagnosing squat issues.

| Predict / Actual | Label 1 | Label 2 | Label 3 | Label 4 | Label 5 | Label 6 | Label 7 |
|---|---|---|---|---|---|---|---|
| Label 1 | 209 | 10 | 11 | 4 | 0 | 0 | 1 |
| Label 2 | 4 | 88 | 4 | 2 | 1 | 0 | 0 |
| Label 3 | 36 | 2 | 43 | 0 | 0 | 0 | 0 |
| Label 4 | 1 | 0 | 0 | 4 | 0 | 0 | 0 |
| Label 5 | 0 | 0 | 0 | 0 | 3 | 0 | 0 |
| Label 6 | 1 | 0 | 0 | 0 | 0 | 11 | 0 |
| Label 7 | 1 | 0 | 0 | 0 | 0 | 0 | 13 |
| F1-Score (Experiment) | 85.83% | 88.44% | 61.87% | 53.33% | 85.71% | 95.65% | 92.85% |
| F1-Score (ML models) | 87.37% | 86.86% | 69.01% | 77.42% | 90.03% | 95.83% | 100% |

**Fig. 11.** Confusion matrix compares prediction results utilizing proposed system (predict) to ground truth based on the opinion of the powerlifting coach (actual). The last two columns are the F1 scores of the experimental results and the testing sets of the selected squat diagnosis models for each label.

## 4 Conclusion

This study successfully developed an intelligent barbell squat coaching system to provide beginners with instant feedback during training. It was the first demonstration of diagnosing multiple squat issues using time series data extracted from the video. Initially, to simplify the input of the predicting models, we created 12 datasets by selecting four critical features for evaluating the barbell squat and three data preprocessing methods. Second, three machine learning architectures, including 1D-CNN, LSTM, and random forest, were adopted to train the diagnosis models to identify six common issues during barbell squats. Subsequently, the SHAP method was employed to optimize feature selection, thereby enhancing the accuracy and computational efficiency of the machine learning models. This study conducted a comparative study involving two groups of participants to validate the system's efficacy: one trained with the proposed system and the other without. Finally, subjects trained with our system outperformed another group with a statistically significant difference. Due to the weight plate hindering the observance of human joints, the proposed system currently couldn't work well when the diameter of the plate was larger than 26 cm. Hence, introducing more filming perspectives while keeping the responding speed will be the topic of further work.

## Funding

This work was funded by the National Science and Technology Council from Taiwan (funding project no. NSTC 113-2410-H-006-128 –, NSTC 113-2425-H-006-006 –) for funding this project.

## Acknowledgements

The authors would like to acknowledge the field support from National Cheng Kung University's Physical Education Office (Chief: Hsien Che Huang).

**References**

1. C. T. Kulik, S. Ryan, S. Harper and G. George, "Aging Populations and Management," *Acad. Manag. J.*, vol.57, no. 4, Jul, 2014, DOI: https://doi.org/10.5465/amj.2014.4004
2. World Population Prospects, "Department of Economic and Social Affairs Population Dynamics," New York: United Nations, 2019, [Online] Available: https://population.un.org/wpp/Download/, Accessed on: Sep. 28, 2020
3. N. Priyanka. N, "Interrelation Between Aging and Sports," [Online] Available: https://www.icliniq.com/articles/healthy-living-wellness-and-prevention/interrelation-between-aging-and-sports, Accessed on: Sep. 17, 2024
4. W. J. Chodzko-Zajko, D. N. Proctor, M. A. Fiatarone Singh, C. T. Minson, C. R. Nigg, G. J. Salem, and J. S. Skinner, "Exercise and Physical Activity for Older Adults," *Med. Sci. Sports Exerc.*, vol.47, no.7, pp. 1510-1530, July 2009, DOI: 10.1249/MSS.0b013e3181a0c95c
5. R. J. Zacker, "Strength Training in Diabetes Management," *Diabetes Spectrum*, vol.18, no. 2, pp. 71-75, 2005, DOI: 10.2337/diaspect.18.2.71
6. F. Mayer, F. Scharhag-Rosenberger, A. Carlsohn, M. Cassel, S. Müller, and J. Scharhag, "The intensity and effects of strength training in the elderly," *Dtsch Arztebl Int.*, May 2011, vol.108, no.21, pp. 359-364, DOI:10.3238/arztebl.2011.0359
7. K. J. Dodd, N. F. Taylor, and S. Bradley, "Chapter 7 - Strength training for older people," in *Optimizing Exercise and Physical Activity in Older People*, 3rd ed. Oxford, United Kingdom: Butterworth-Heinemann, 2004, pp. 143.
8. R. Kostić, S. Uzunovic, S. Pantelić, and R. Đurašković, "A comparative analysis of the indicators of the functional fitness of the elderly," *Facta Universitatis: Series Physical Education & Sport,* 2011, vol.9, pp. 161-171.
9. T. Tsutsumi, B. M. Don, L. D. Zaichkowsky, and L. L. Delizonna, "Physical fitness and psychological benefits of strength training in community dwelling older adults," *Appl Human Sci.*, Nov. 1997, vol.16, no.6, pp. 257-266, DOI:10.2114/jpa.16.257
10. Y. Netz, M.-J. Wu, B. J. Becker, and G. Tenenbaum, "Physical activity and psychological well-being in advanced age: a meta-analysis of intervention studies," *Psychol Aging.*, Jun. 2005, vol.20, no.2, pp. 272-284, DOI:10.1037/0882-7974.20.2.272
11. M. M. Orlando, M. S. P. da Silva, and I. Lombardi Jr., "The influence of the practice of physical activity on the quality of life, muscle strength, balance, and physical ability in the elderly," *Rev. Bras. Geriatr. Rerontol.*, Mar. 2013, vol.16, no.1, pp. 117-126, DOI:10.1590/S1809-98232013000100012
12. E. Carmeli, A. Z. Reznick, R. Coleman, and V. Carmeli, "Muscle strength and mass of lower extremities in relation to functional abilities in elderly adults," *Gerontology*, Sep. 2000, vol.46, no.5, pp. 249-257, DOI:10.1159/000022168
13. M. Chiacchiero, B. Dresely, U. Silva, R. DeLosReyes, and B. Vorik, "The Relationship Between Range of Movement, Flexibility, and Balance in the Elderly," *Top. Geriatr. Rehabil.*, Apr. 2010, vol.26, no.2, pp. 148-155, DOI:10.1097/TGR.0b013e3181e854bc
14. V.J. Wright, and F. Tejpar, "The New Science of Musculoskeletal Aging in Bone, Muscle, and Tendon/Ligament," in *Masterful Care of the Aging Athlete*, Berlin, Germany: Springer, 2018, pp. 9-15, DOI: 10.1007/978-3-319-16223-2_2
15. S. M. Nam, and Y. Eun, "The Influence of Exercise Intention, Exercise Habits, and Exercise Self-efficacy on Regular Exercise Behaviors in the Elderly," *J Muscle Jt Health.*, Dec. 2019, vol.26, no.3, pp. 232-240, DOI:10.5953/JMJH.2019.26.3.232
16. E. M. Phillips, J. C. Schneider, and G. R. Mercer, "Motivating elders to initiate and maintain exercise," *Arch Phys Med Rehabil.*, Jul. 2004, vol.85, no.7 Suppl 3, pp. S52-57, DOI:10.1016/j.apmr.2004.03.012
17. F. W. Booth, C. K. Roberts, and M. J. Laye, "Lack of exercise is a major cause of chronic diseases," *Compr Physiol.*, Apr. 2012, vol.2, no.2, pp. 1143-1211, DOI:10.1002/cphy.c110025
18. E. Anderson, and J. L. Durstine, "Physical activity, exercise, and chronic diseases: A brief review," *Sports Med Health Sci.*, Dec. 2019, vol.1, no.1, pp. 3-10, DOI:10.1016/j.smhs.2019.08.006
19. E. Saraee, S. Singh, A. Joshi, and M. Betke, "PostureCheck: Posture modeling for exercise assessment using the Microsoft Kinect," in Proc. *ICVR*, Montreal, QC, Canada, 2017, pp. 1-2, DOI:10.1109/ICVR.2017.8007497
20. L. Grewe, D. N. T. Pham, D. Pravin Jain, A. Mahajan, and A. Shahshahani, "A detective and corrective exercise assistant using computer vision and machine learning," in Proc. *SPIE 12122, Signal Processing, Sensor/Information Fusion, and Target Recognition XXXI, 121220X*, Orlando, Florida, United States, 2022, DOI:10.1117/12.2619102
21. T. T. Than, D. K. D. Danh, H. L. Nguyen, and M. S. Nguyen, "Researching and Implementing the Posture Recognition Algorithm of the Elderly on Jetson Nano," in Proc. *International Conference on Multimedia Analysis and Pattern Recognition (MAPR)*, Phu Quoc, Vietnam, 2022, pp. 1-5, DOI:10.1109/MAPR56351.2022.9924968
22. S. Ghazal, U. S. Khan, M. M. Saleem, N. Rashid, and J. Iqbal, "Human activity recognition using 2D skeleton data and supervised machine learning," *IET Image Processing*, Nov. 2019, vol.13, no.13, pp. 2572-2578, DOI:10.1049/iet-ipr.2019.0030
23. K.-Y. Chen, J. Shin, M. A. M. Hasan, J.-J. Liaw, O. Yuichi, and Y. Tomioka, "Fitness Movement Types and Completeness Detection Using a Transfer-Learning-Based Deep Neural Network," *Sensors*, Jul. 2017, vol.22, no.15, pp. 5700, DOI:10.3390/s22155700

24. G. Zhu, L. Zhang, P. Shen, and J. Song, "An Online Continuous Human Action Recognition Algorithm Based on the Kinect Sensor," *Sensors*, Jan 2016, vol.16, no.2, pp. 161, DOI:10.3390/s16020161
25. A. Gupta, K. Gupta, K. Gupta, and K. O. Gupta, "Human Activity Recognition Using Pose Estimation and Machine Learning Algorithm," in Proc. *International Semantic Intelligence Conference*, New Delhi, India, 2017, pp. 323-330
26. S. M. Lundberg, and S.-I. Lee, "A Unified Approach to Interpreting Model Predictions," in Proc. *Advances in Neural Information Processing Systems 30*, pp. 4768–4777, DOI:10.48550/arXiv.1705.07874